\documentclass[letterpaper]{article} 
\usepackage{aaai2026}  
\usepackage{times}  
\usepackage{helvet}  
\usepackage{courier}  
\usepackage[hyphens]{url}  
\usepackage{graphicx} 
\usepackage{natbib}  
\usepackage{caption} 
\usepackage{algorithm}
\usepackage{algorithmic}

\usepackage{amsmath}
\usepackage{amsfonts}
\usepackage{amssymb}
\usepackage{mathrsfs}
\usepackage{multirow}
\usepackage{booktabs} 
\usepackage{newfloat}
\usepackage{listings}
\DeclareCaptionStyle{ruled}{labelfont=normalfont,labelsep=colon,strut=off} 
\floatstyle{ruled}
\newfloat{listing}{tb}{lst}{}
\floatname{listing}{Listing}
\nocopyright

\title{Spike Imaging Velocimetry:\\Dense Motion Estimation of Fluids Using Spike Streams}
\author{
    Yunzhong Zhang\textsuperscript{\rm 1,\rm 2}\equalcontrib, You Zhou\textsuperscript{\rm 3}\equalcontrib, Changqing Su\textsuperscript{\rm 4}, Zhen Cheng\textsuperscript{\rm 5}, Zhaofei Yu\textsuperscript{\rm 4}, Bo Xiong\textsuperscript{\rm 6}\thanks{Corresponding author.}, Tiejun Huang\textsuperscript{\rm 4}, Xun Cao\textsuperscript{\rm 2}
}
\affiliations{
    \textsuperscript{\rm 1}School of Physics, Nanjing University\\
    \textsuperscript{\rm 2}School of Electronic Science and Engineering, Nanjing University\\
    \textsuperscript{\rm 3}Medical School, Nanjing University\\
    \textsuperscript{\rm 4}State Key Laboratory for Multimedia Information Processing, Peking University\\
    \textsuperscript{\rm 5}Department of Automation, Tsinghua University\\
    \textsuperscript{\rm 6}Institute of Medical Technology, Peking University Health Science Center, Peking University\\

    ltq@smail.nju.edu.cn, \{zhouyou, caoxun\}@nju.edu.cn, cqsu25@stu.pku.edu.cn, zcheng@mail.tsinghua.edu.cn, \{yuzf12, xiongbo, tjhuang\}@pku.edu.cn, 
}

\usepackage{bibentry}

\begin{document}

\maketitle


\begin{abstract}
Particle Image Velocimetry (PIV) is a widely adopted non-invasive imaging technique that tracks the motion of tracer particles across image sequences to capture the velocity distribution of fluid flows. It is commonly employed to analyze complex flow structures and validate numerical simulations. This study explores the untapped potential of spike cameras—ultra-high-speed, high-dynamic-range vision sensors—in high-speed fluid velocimetry. We propose a deep learning framework, Spike Imaging Velocimetry (SIV), tailored for high-resolution fluid motion estimation. To enhance the network’s performance, we design three novel modules specifically adapted to the characteristics of fluid dynamics and spike streams: the Detail-Preserving Hierarchical Transform (DPHT), the Graph Encoder (GE), and the Multi-scale Velocity Refinement (MSVR). Furthermore, we introduce a spike-based PIV dataset, Particle Scenes with Spike and Displacement (PSSD), which contains labeled samples from three representative fluid-dynamics scenarios: steady turbulence, high-speed flow, and high-dynamic-range conditions. Our proposed method outperforms existing baselines across all these scenarios, demonstrating its effectiveness.
\end{abstract}

\section{Introduction}
\begin{figure}[ht]
  \centering
  \resizebox{1\linewidth}{!}{%
    \includegraphics{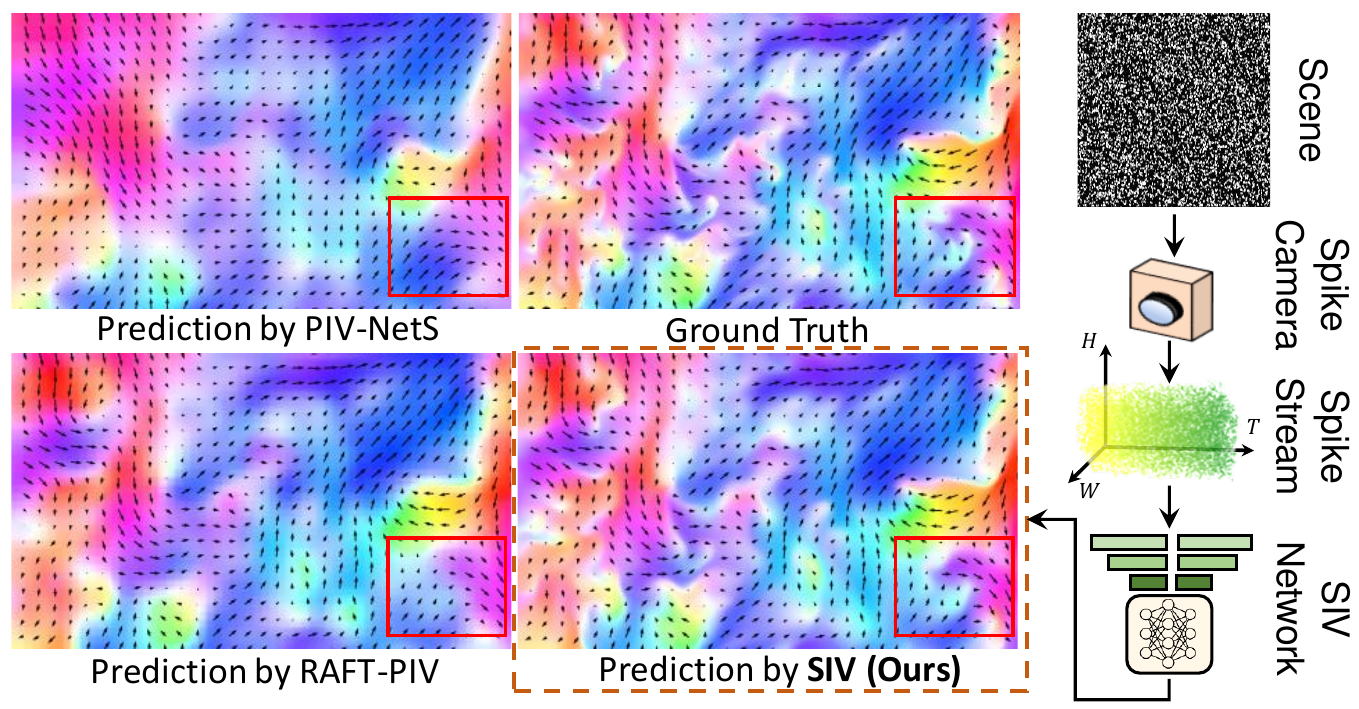}
  }
  \caption{A schematic of fluid velocity field estimation based on spike cameras, and a comparison of the complex velocity field estimated using our proposed SIV network with PIV-NetS \protect\cite{pivnets} and RAFT-PIV \protect\cite{raftpiv}.}
  \label{introduction}
\end{figure}
The need for accurate and non-invasive flow measurement techniques has led to the widespread use of Particle Image Velocimetry (PIV), a key diagnostic tool for fluid dynamics. PIV has been applied to critical areas such as studying aircraft aerodynamics \cite{kompenhans2000particle}, optimizing vehicle designs for reduced air resistance \cite{nakagawa2016typical}, improving energy efficiency in combustion processes \cite{chen2022investigations}, understanding blood flow dynamics in diseases \cite{raschi2012cfd}, and predicting water pollution spread \cite{moayerikashani2016tracking}. In PIV experiments, tracer particles scatter light within a fluid, and their displacements are used to infer the velocity field of the fluid.
Recent advancements in optical flow networks \cite{flownet,pwcnet,raft} have inspired the development of learning-based PIV algorithms \cite{pivnets,pivliteflownet,raftpiv}. While optical flow networks are highly generalizable, fluid scenes present unique challenges that limit further performance improvements. Turbulence, a chaotic flow phenomenon at high Reynolds numbers, introduces specific issues in velocity field estimation: \textbf{(1)} High-speed fluid movement within the field of view. \
\textbf{(2)} Sparse regions without tracer particles, leading to discrete signals while still requiring the estimation of a continuous flow field. \
\textbf{(3)} Unstructured patterns and shapes with small-scale vortices and quasi-ordered structures. These challenges coexist in turbulent flows, necessitating advanced experimental hardware and algorithmic solutions. To address these, we propose Spike Imaging Velocimetry (SIV). In terms of hardware, we explore the use of spike cameras \cite{firstspike,spikepatent} for PIV. Spike cameras offer several advantages for fluid velocity measurement: \\ - \noindent \textbf{High Temporal Resolution (HTR):} With a temporal resolution of 20,000 Hz, spike cameras significantly reduce the time interval $\Delta t$ between consecutive frames, providing rich temporal information that enables precise predictions. This addresses Challenge (1) effectively. \\ - \noindent \textbf{High Dynamic Range (HDR):} Spike cameras asynchronously accumulate photons at each pixel, providing a larger dynamic range, which is beneficial in scenarios with uneven lighting or high particle concentration. This helps mitigate limitations seen in conventional cameras, addressing Challenge (2). \\ - \noindent \textbf{Reduced Experimental Complexity:} Spike cameras can continuously capture high-speed images with low bandwidth and do not require synchronization controllers, simplifying experimental setups. \\

We propose a specialized network, \textbf{Spike Imaging Velocimetry (SIV)}, for estimating fluid velocity fields. To enhance network performance, we develop three novel modules tailored to the characteristics of fluid dynamics and spike cameras. These include the \textbf{Detail-Preserving Hierarchical Transform (DPHT)}, which enhances feature extraction from spike streams while mitigating downsampling artifacts; the \textbf{Graph Encoder (GE)}, which maps features into a graph structure for adaptive context aggregation; and the \textbf{Multi-scale Velocity Refinement (MSVR)}, which addresses over-smoothing in predicted flow fields. We also introduce a new dataset, \textbf{Particle Scenes with Spike and Displacement (PSSD)}, designed for learning fluid motion from spike data. It covers three representative motion estimation scenarios: steady turbulence, high-speed flow, and high dynamic range conditions. Extensive experiments demonstrate the superiority of SIV over both image-based and spike-based baselines. SIV achieves state-of-the-art performance across all evaluation settings. The effectiveness of each proposed module is further validated through ablation studies. The contributions of this work are summarized as follows:\\
\noindent \textbf{1.} We explore a feasible solution for applying spike cameras to fluid velocity measurement, opening new directions for PIV.\\
\noindent \textbf{2.} We design a dedicated network, SIV, tailored to fluid dynamics and turbulent flow properties, and propose three novel modules: the DPHT, GE, and MSVR.\\
\noindent \textbf{3.} We propose a new dataset, PSSD, which captures continuous fluid motion across four flow types and three challenging experimental scenarios.\\
\noindent \textbf{4.} We conduct extensive quantitative experiments and qualitative visual comparisons, demonstrating that SIV achieves state-of-the-art performance across multiple benchmarks. We further perform ablation studies to systematically validate the effectiveness of each proposed module.

\section{Related Work}
 
\subsection{Spike-based Optical Flow Estimation}
The spike camera \cite{firstspike,xiong2024real,su2024intensity} is a type of neuromorphic camera (NeurCam). Inspired by biological visual neural structures and mechanisms, NeurCams emulate the encoding principles of the primate retina. Spike cameras mimic the encoding principles of the central fovea in the retina, providing continuous visual information with ultra-high temporal resolutions, high dynamic range (HDR), and significantly reduced energy consumption and latency, making them well-suited for high-speed motion scenarios. Their pixels independently accumulate photons and emit a spike when the accumulated intensity exceeds a certain threshold. This process maps the incoming photon stream into the spike stream, which preserves spatiotemporal relationships and can be reconstructed into images. For a detailed explanation of the working mechanism, refer to \cite{firstspike}. The application of spike cameras in optical flow estimation provides significant insights for this study. Although optical flow estimation is generally used to estimate the motion of macroscopic objects, its principles and objectives are aligned with those of PIV. The techniques employed in PIV for image registration and displacement estimation are essentially a variant of optical flow estimation. Spike-based optical flow estimation is still a relatively new field. \cite{hu2022optical} introduced a specialized neural network architecture for spike-based optical flow estimation. This structure can adaptively eliminate motion blur in spike streams based on prior motion information. A framework based on RAFT is proposed by \cite{spike2flow} to handle continuous spike streams. \cite{hist} demonstrates that using a multi-layer hierarchical structure to characterize spike streams can effectively suppress the influence of randomness in spikes and enhance the feature extraction capability of the network. \cite{chen2023self} proposed a self-supervised approach to learn optical flow from spike cameras, which is then extended by \cite{xia2024unsupervised}.

\subsection{Image-based PIV Algorithms}

Many traditional image-based PIV algorithms \cite{classicpiv1,classicpiv3} usually involve applying a fast Fourier transform (FFT) between two subregions of image pairs, known as interrogation windows. Displacement is then determined by finding the maximum correlation between the two interrogation windows, which is similar to the correlation method in optical flow estimation to some extent. However, traditional methods depend on finely tuned parameters and complex optimization techniques. With advancements in optical flow networks, various architectures \cite{flownet,pwcnet,raft} have been developed to predict dense velocity fields of macroscopic objects with high accuracy. Early PIV networks like PIV-DCNN \cite{pivdcnn} laid the groundwork for these developments. PIV-NetS \cite{pivnets} is capable of generating full-resolution displacement outputs. \cite{pivliteflownet} adapt LiteFlowNet \cite{liteflownet} to PIV. RAFT-PIV \cite{raftpiv}, derived from RAFT \cite{raft}, has shown exceptional performance on multiple datasets. \cite{lightpivnet} removes the computationally inefficient components of RAFT, such as the context encoder, to improve inference speed. Additionally, unsupervised learning frameworks for PIV\cite{unsupervisedpiv1,unsupervisedpiv2} have been introduced, reducing reliance on the extremely difficult-to-obtain ground-truth displacements. Other recent PIV networks include LIMA \cite{manickathan2023lightweight}, CC-FCN \cite{gao2021robust}, En-FlowNetC \cite{ji2024cross}, ARAFT-FlowNet \cite{han2023attention}, OFVNetS \cite{ji2024effects}, and PIV-PWCNet \cite{zhang2023pyramidal}. However, the performance gains of most PIV networks are largely attributed to advances in optical flow estimation architectures, rather than targeted enhancements tailored to the specific characteristics of fluid dynamics, and most implementations remain unavailable as open-source.

\subsection{Event-based PIV Algorithms}
The event camera \cite{dvs,davis,atis,jiang2024edformer} is another type of NeurCam, which is inspired by the peripheral retina \cite{event1}. Unlike spike cameras, event cameras independently monitor the change in light intensity (increase or decrease) at each pixel and generate signals based on the local pixel information of dynamic changes. When the intensity change of a pixel exceeds a preset threshold, the camera records this change event (intensity increase as a positive event, intensity decrease as a negative event) and outputs its timestamp, position, and change value as data. The success of event cameras in optical flow estimation \cite{lee2020spike-flownet,shiba2022secrets} has paved the way for their application to PIV. \cite{EBIV} introduced the first event-camera PIV algorithm based on correlation and motion compensation. \cite{eventpiv} combined an event camera with pulsed illumination, substantially reducing estimation latency. These studies are insightful and practical; however, their algorithmic novelty is limited, and robustness is partly constrained by the event camera’s lack of absolute intensity.

\section{Preliminary}

\subsection{The Working Principle of the Spike Camera}

Each photosensitive element in a spike camera generates spike streams sequentially over time. A "1" in the stream indicates that a spike has occurred (i.e., the threshold has been reached), while a "0" indicates an ongoing accumulation state. The Integral-and-Fire (IF) mechanism of spike cameras can be mathematically represented as follows:
\begin{equation}
A(\mathbf{x},t)=\left( \int_{0}^{t}\alpha \cdot I(\mathbf{x},\tau)d\tau \right) \bmod \theta,
\end{equation}

Where $A(\mathbf{x},t)$ represents the voltage of the accumulator at pixel $\mathbf{x}=(x,y)$ at time $\tau$, and $\theta$ is the threshold of the comparator. The spike camera's output is a spatiotemporal binary stream $S$ of size $H \times W \times N$, where $H$ and $W$ are the height and width of the sensor, respectively, and $N$ represents the temporal dimension. Compared to traditional cameras, spike cameras preserve richer visual information regarding both temporal resolution and intensity representation.

\subsection{Data Generation}
Training neural networks using supervised strategies requires data with ground truth to optimize model parameters. However, it is difficult to obtain accurate velocity fields from experimental PIV data. We have to generate a synthetic dataset for training, following \cite{pivnets}. Our proposed dataset, Particle Scenes with Spike and Displacement (PSSD), includes three sub-datasets: Steady Turbulence (Problem 1), High-speed Flow (Problem 2), and HDR Scenes (Problem 3). Each sub-dataset comprises 10,000 sample sequences of spike streams and the corresponding ground-truth turbulent velocity fields when $\Delta t = 21$ (approximately 1.05 ms) and $\Delta t = 11$ (approximately 0.55 ms). Additionally, we provide corresponding pairs of images for comparative analysis for image-based methods. We use the reference flow field to simulate the motion of the scene and construct a virtual spike camera to trigger the spike stream. Both the images and the spike stream in the dataset include added noise. The reference flow fields are obtained from the Johns Hopkins Turbulence Database (JHTDB) \cite{JHTDB}, a high-precision numerical simulation database widely used in turbulence research. JHTDB is based on direct numerical simulation (DNS) of the Navier–Stokes equations without any turbulence modeling, resolving all dynamically relevant scales in both space and time on fine grids. As a result, the simulated flow fields closely approximate real turbulent flows and can be regarded as ground truth in our study.

\begin{figure}[ht]
  \centering
  \includegraphics[width=\linewidth]{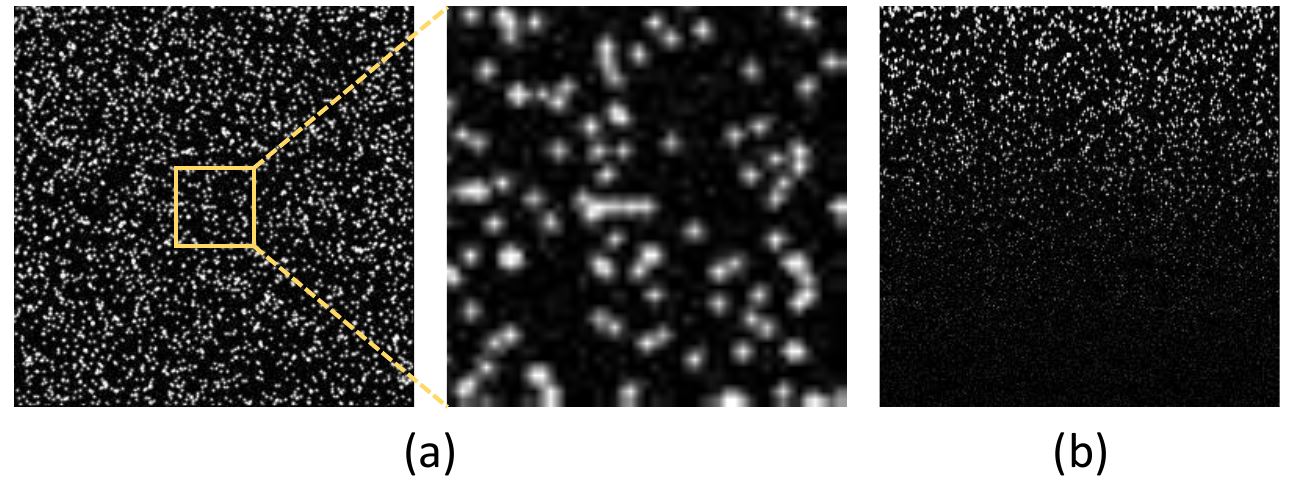}
  \caption{(a) Normal illumination scene (Problem 1, Problem 2); (b) High dynamic range scene (Problem 3).}
  \label{fig:piv_scene}
\end{figure}

\section{Approaches}

\subsection{Overall Architecture}
\noindent \textbf{Problem definition.} Let $\mathbf{S} \in \mathbb{B}^{H\times W \times T}$ represent the spike stream recorded by the spike camera. $\mathbb{B}$ is the binary domain. We aim to estimate the displacement $\mathbf{D}$ of particles in the fluid between time moments $t$ and $t+\Delta t$. At the pixel level, for any given pixel $\mathbf{p}=(x,y)$ at $t$, we need to find a unique correspondence formulated by:
\begin{equation}
    \mathcal{I}(\mathbf{p},t) \leftarrow \mathcal{I}(\mathbf{p}+\mathbf{D}(\mathbf{p}),t+\Delta t).
\end{equation}
Here, $\mathcal{I}$ is the mapping of the spike stream $\mathbf{S}$ in the high-dimensional feature space, representing the scene information.

\begin{figure}[ht]
  \centering
    \includegraphics[width=1.0\linewidth]{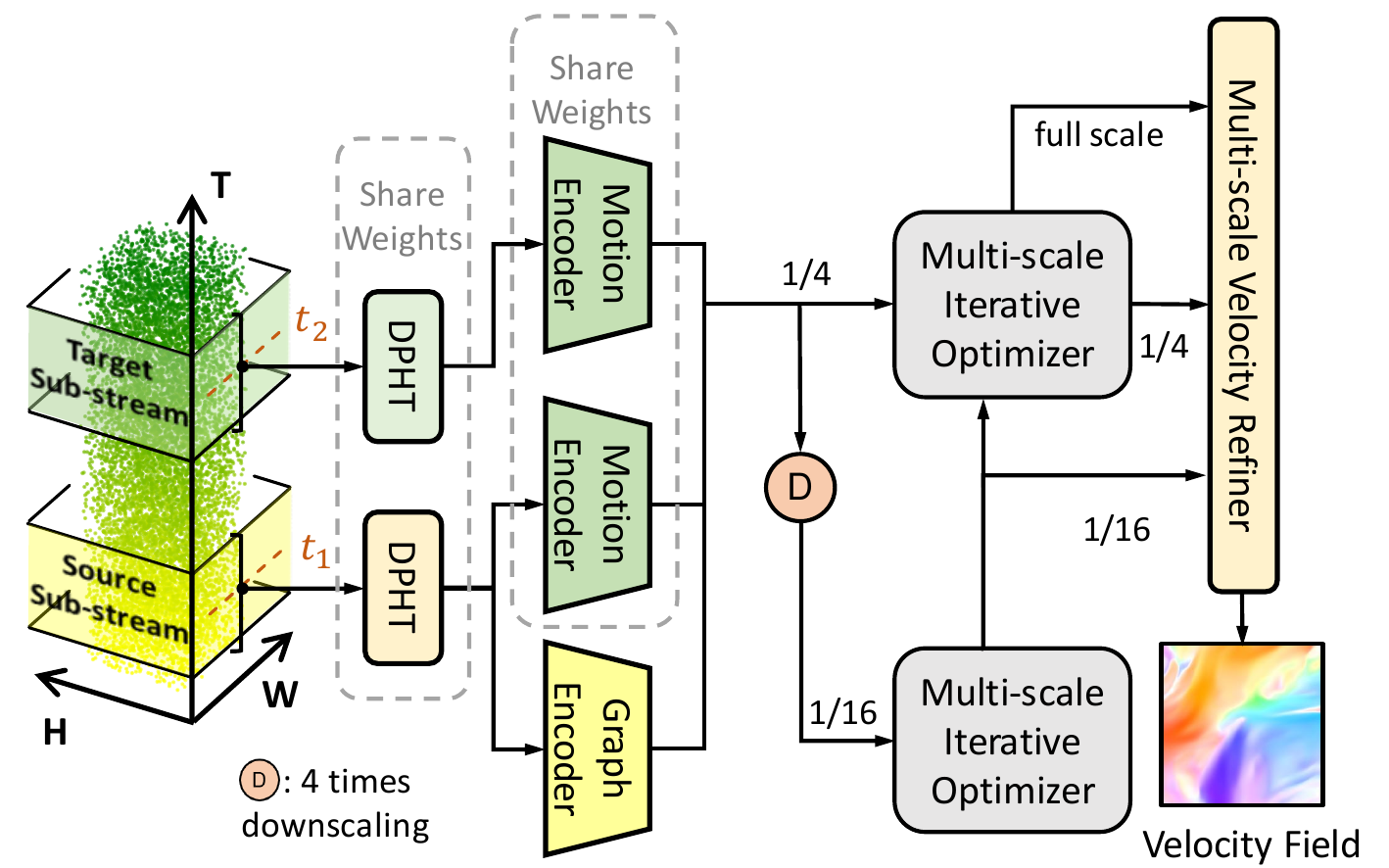}
  \caption{
  Spike Imaging Velocimetry (SIV) network.
}
\label{backbone}
\end{figure}

\begin{figure*}[t]
  \centering
    \includegraphics[width=1.0\linewidth]{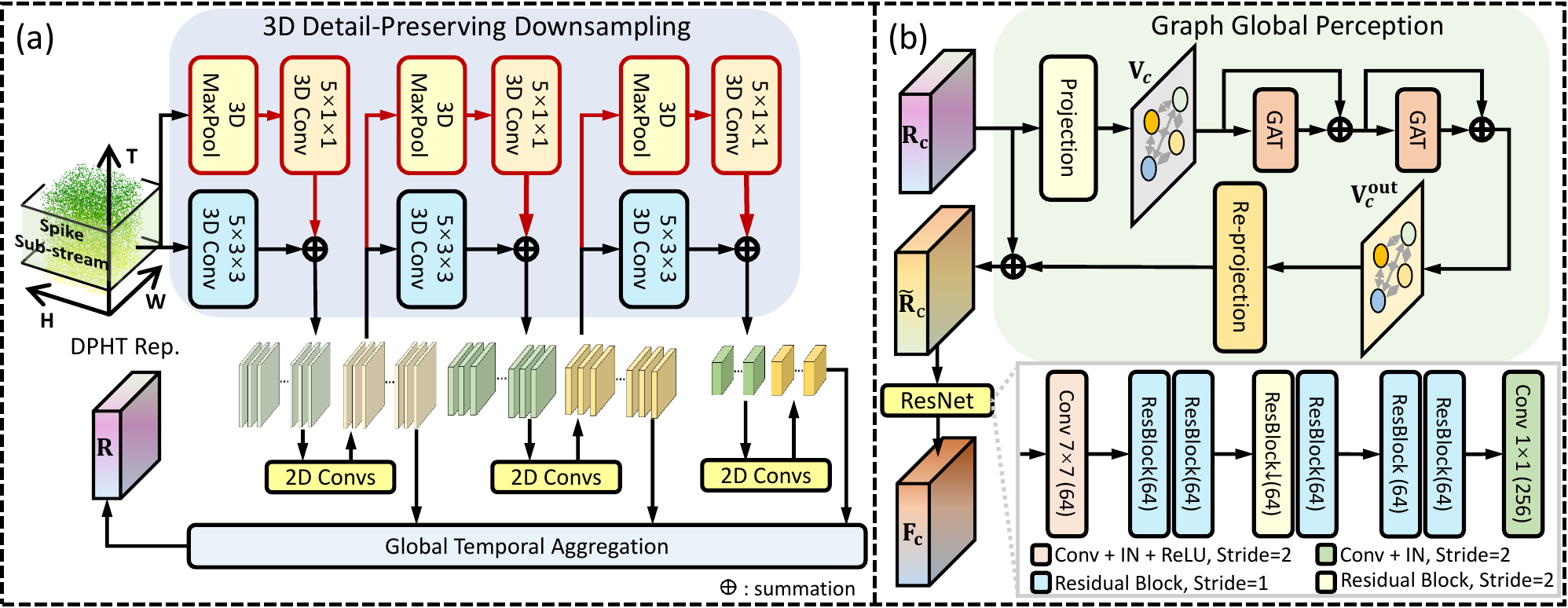}
  \caption{
  (a) Detail-Preserving Hierarchical Transform (DPHT) Module. (b) Graph Encoder (GE).
  }
\label{modules}
\end{figure*}

\subsection{Detail-Preserving Hierarchical Transform}

To preserve particle signal information during feature extraction, we propose the Detail-Preserving Hierarchical Transform (DPHT), a multi-scale pyramidal architecture that integrates a 3D Detail-Preserving Downsampling module (see Figure \ref{modules}(a)).

\noindent \textbf{3D Detail-Preserving Downsampling.} Given an input feature vector $\mathbf{x}^{(l)}$ at level $l$, the processing is performed as follows:
\begin{equation}
\mathbf{x}^{(l)}_{\mathrm{out}} = \mathscr{D}^{(l)}(\mathbf{x}^{(l)})= \mathscr{F}_{\mathrm{2C}}(\mathscr{F}_{\mathrm{3C}}(\mathbf{x}^{(l)}) + \mathscr{F}_{\mathrm{3MP}}(\mathbf{x}^{(l)})),
\end{equation}
where $\mathscr{F}_{\mathrm{3C}}$ is a 5×3×3 3D convolution, $\mathscr{F}_{\mathrm{3MP}}$ combines 3D max pooling and a 5×1×1 3D convolution, and $\mathscr{F}_{\mathrm{2C}}$ refers to a stack of 2D convolutions. The output of each level is passed to the next, i.e., $\mathbf{x}^{(l+1)} = \mathbf{x}^{(l)}_{\mathrm{out}}$.

\noindent \textbf{Global Temporal Aggregation.} The output feature vectors from each level are refined through additional convolutions and then concatenated to form the final DPHT representation:
\begin{equation}
\mathbf{R} = \mathscr{A}{m}[\mathbf{Concat}(\mathbf{x}^{(l=1)}_{\mathrm{out}}, \mathbf{x}^{(l=2)}_{\mathrm{out}}, \mathbf{x}^{(l=3)}_{\mathrm{out}})],
\end{equation}
where $\mathscr{A}_{m}$ denotes a convolution operation. The source and target sub-streams' signals are represented as $\mathbf{R}_s$ and $\mathbf{R}_t$, respectively.

\subsection{Graph Encoder}

In turbulent flows with high Reynolds numbers, large vortices break into smaller ones, revealing an underlying topological structure in fluid motion. To capture this structure, we propose the Graph Encoder (GE), a feature extraction network based on Graph Neural Networks (GNNs) and Convolutional Neural Networks (CNNs). By representing features as graph nodes and constructing a dynamic adjacency matrix, GE aggregates and updates information from neighboring nodes, allowing for a larger receptive field than traditional CNNs. Graph Attention Networks (GATs) \cite{GAT} leverage attention mechanisms to learn the relative contribution of each neighbor, reducing interference from particle-sparse regions. This combination of GNN and CNN enables adaptive global perception and local refinement, effectively capturing fluid-specific contextual features.

\noindent \textbf{Graph Projection.} The first step is to map $\mathbf{R}_s$ from the DPHT to the node space, i.e., $\mathbf{V}_{c}=\mathscr{P}_{R \to V}(\mathbf{R}_{s})$. The context node representations, $\mathbf{V}_{c} = \{v_{c}^{1},\cdots,v_{c}^{n}\} \in \mathbb{R}^{C \times K}$, contain appearance features related to the scene’s shape and regional information, where $C=128$ is the channel number and $K$ is the number of nodes (set to $K=128$ in this paper). The projection is formulated as:
\begin{equation}
v_{c}^{i} = \mathscr{N}(\sum\limits_{\forall{j}}z_{ij} \cdot \mathbf{R}_{s,j}),
\end{equation}
where $\mathscr{N}(\cdot)$ is the $L_2$ normalization, and $z_{ij}$ denotes the normalized projection weights computed by a $1\times1$ convolution and a softmax function.

\noindent \textbf{Graph Convolution Layer.} After obtaining the nodes $\mathbf{v}_c$, the dynamic adjacency matrix $\widetilde{\mathbf{A}} = \mathbf{V}_{c}^{T} \mathbf{V}_{c}$ is computed based on the similarity between node vectors. The operations of two GAT modules, $\mathscr{G}_1$ and $\mathscr{G}_2$, update the node features as follows:
\begin{equation}
\mathbf{V_{c}^\mathrm{out}} = \mathscr{G}_2[\mathscr{G}_1(\mathbf{V}_{c}, \widetilde{\mathbf{A}}) + \mathbf{V}_{c}, \widetilde{\mathbf{A}}] + \mathscr{G}_1(\mathbf{V}_{c}, \widetilde{\mathbf{A}}) + \mathbf{V}_{c},
\end{equation}

\noindent \textbf{Graph Re-projection.} The re-projection operation $\mathscr{P}_{V \to R}(\cdot)$ takes the transformed vertex features $\mathbf{V_{c}^\mathrm{out}}$ and the projection weight matrix $\mathbf{Z}$, producing a 2D feature map $\widetilde{\mathbf{R}}c$:
\begin{equation}
\widetilde{\mathbf{R}}_c = \mathscr{P}_{V \to R}(\mathbf{V_{c}^\mathrm{out}}, \mathbf{Z}) = \mathbf{Z} \left( \mathbf{V_{c}^\mathrm{out}} \right)^{T}.
\end{equation}
This operation preserves spatial details, as even if two pixels share the same node, they will have distinct features. Finally, these results are integrated into a residual extractor network $\mathscr{R}$ for local refinement:
\begin{equation}
\mathbf{F}_c = \mathscr{R}(\mathbf{R}_s + \alpha \widetilde{\mathbf{R}}_c),
\end{equation}
where $\alpha$ is a learnable parameter initialized to 0, gradually refining the feature sum.


\subsection{Multi-scale Iterative Optimizer}

We adopt the Multi-scale Iterative Optimizer (MSIO) from DIP \cite{dip} as the backbone for iterative residual flow estimation in SIV. Unlike RAFT’s full correlation volume construction, MSIO leverages Inverse PatchMatch to significantly reduce the computational cost introduced by large-scale feature maps (1/4 resolution of the full size). The detailed architecture of MSIO can be found in \cite{dip}; here, we denote it implicitly as $\mathscr{U}$. The output multi-scale velocity field is given by
\begin{equation}
\{\mathbf{u}_i\}_{i=1}^{N} = \mathscr{U}(\mathbf{F}_c, \mathbf{F}_m),
\end{equation}
where $\mathbf{F}_m$ denotes motion features extracted by the Motion Encoder, and $N$ refers to the number of GRU iterations \cite{raft}.

\subsection{Multi-scale Velocity Refinement}

\begin{figure}[ht]
  \centering
    \includegraphics[width=\linewidth]{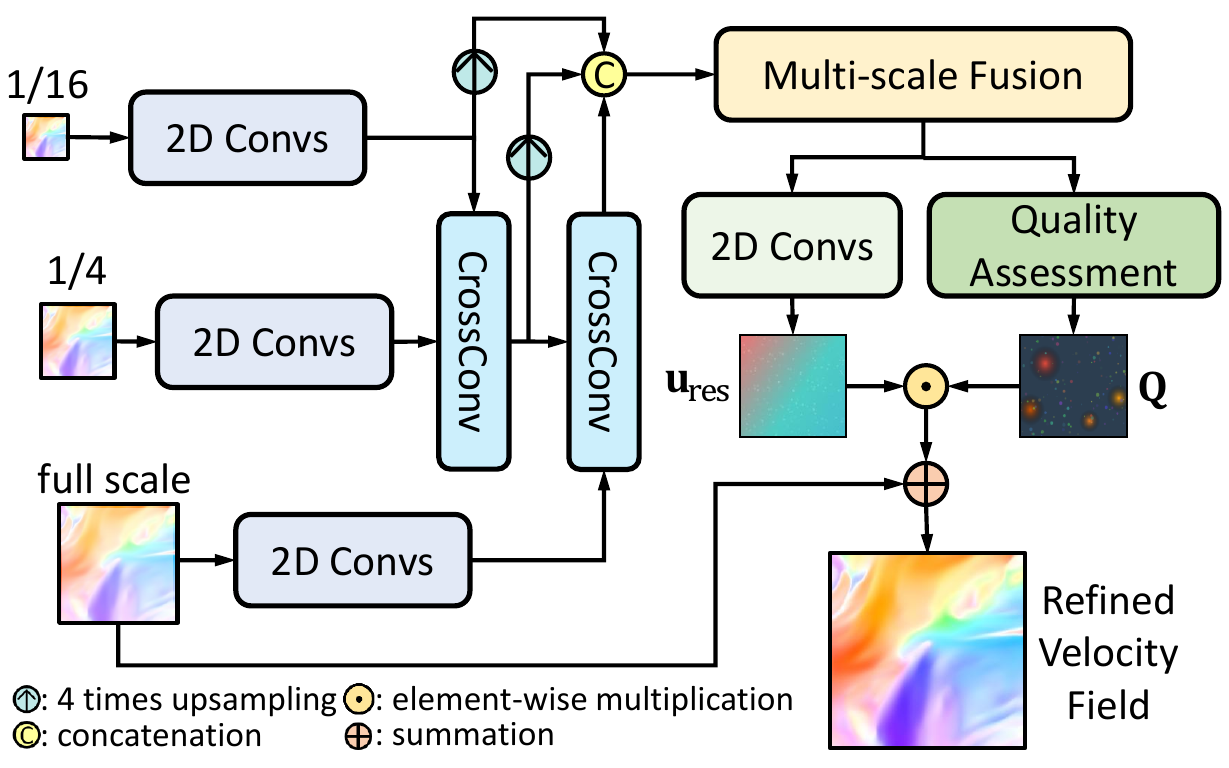}
  \caption{
  Multi-scale Velocity Refinement Module.
}
\label{module_c}
\end{figure}

DIP adopts the same compact and effective convex upsampling strategy as RAFT to recover full-resolution optical flow. However, such upsampling is insufficient for accurately reconstructing the velocity field of fluids. To address this, we propose a Multi-scale Velocity Refinement (MSVR) module to assist in reconstructing the complete velocity field, including fine-scale vortices. As illustrated in Figure \ref{module_c}, we first apply a sequence of 2D convolutions followed by activation functions to process the multi-scale velocity fields generated by MSIO. Information exchange across scales is achieved via cross convolution from coarse to fine levels. The extracted features are then fused and processed by distinct convolutional networks to generate a residual velocity field  $\mathbf{u}_{\text{res}}$ and a quality map $\mathbf{Q}$. The refined fluid velocity field is given by,

\begin{equation}
\mathbf{u}_\text{ref} = \mathbf{u}_N + \mathbf{u}_{\text{res}} \odot \mathbf{Q},
\end{equation}
where $\odot$ denotes element-wise multiplication.

\subsection{Loss Function}
The loss function for the proposed network is composed of two parts: the flow loss and the gradient loss. Suppose the recurrent optimizer of the network has $N$ iterations, and the estimated flow fields of each iteration are $\{\mathbf{u}_1,\dots,\mathbf{u}_N,\mathbf{u}_{N+1}\}$, $\mathbf{u}_{N+1}=\mathbf{u}_\text{ref}$.
The flow loss is defined as:
\begin{equation}
L_\text{flow} = \sum_{i=1}^{N+1} \gamma^{N+1-i} \cdot \|\mathbf{u}_i - \mathbf{u}_\text{gt}\|_1,
\end{equation}
$\mathbf{u}_\text{gt}$ is the ground truth of velocity field. To preserve small vortical structures, we introduce a gradient loss:
\begin{equation}
\begin{aligned}
    L_\text{grad} &= \sum_{i=1}^{N+1} \gamma^{N+1-i} \cdot (\|\nabla_{\mathbf{x}} \mathbf{u}_i - \nabla_{\mathbf{x}} \mathbf{u}_\text{gt}\|_1 \\
    &+ \|\nabla_{\mathbf{y}} \mathbf{u}_i - \nabla_{\mathbf{y}} \mathbf{u}_\text{gt}\|_1). 
\end{aligned}
\end{equation}
The total loss is:
\begin{equation}
L = L_\text{flow} + \beta \cdot L_\text{grad}.
\end{equation}
$\gamma$ is set to 0.8, and $\beta$ is set to 0.3.

\section{Experiments}
We evaluate our method on the PSSD dataset and compare its performance with both traditional and learning-based approaches. Additionally, we conduct a series of ablation studies to assess the effectiveness of the proposed modules.

\begin{figure*}[ht]
  \centering
    \includegraphics[width=1.0\linewidth]{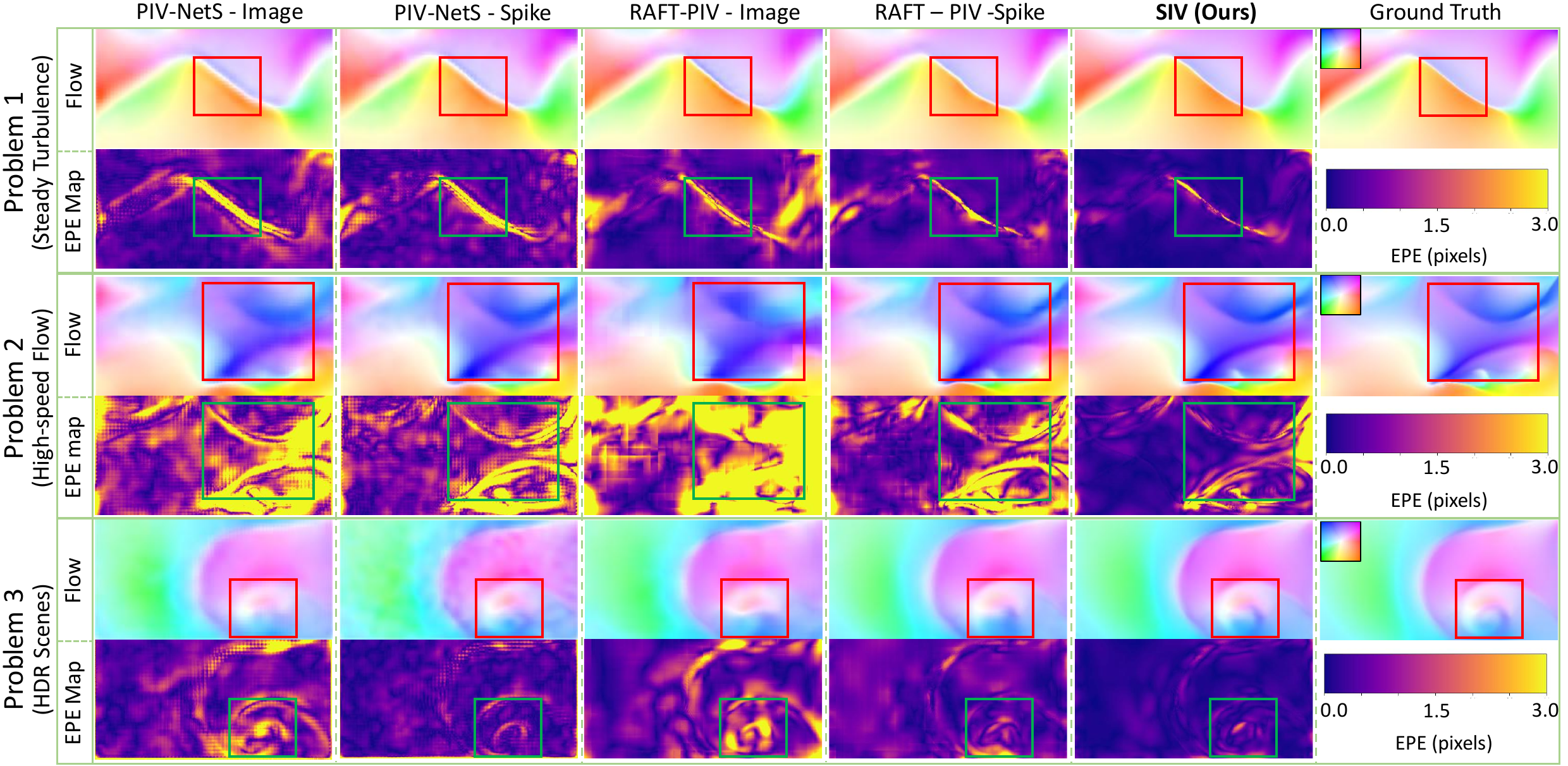}
  \caption{
Visualization examples of different algorithms in Steady Turbulence (Problem 1), High-speed Flow (Problem 2), and HDR Scenes (Problem 3) are shown. For each problem, the red box in the top optical flow map and the green box in the bottom EPE error map indicate the same region of the flow field. 
}
\label{visualization}
\end{figure*}

\subsection{Experimental Details}
SIV is implemented using the PyTorch framework on a single NVIDIA RTX 2080 Ti GPU. We use the Adam optimizer with a warm-up strategy during training. For each sub-dataset, the model is trained for 100 epochs with a batch size of 4 and an initial learning rate of $10^{-4}$, applying an 85\% decay every 10 epochs. The number of iterative refinements for the GRUs is set to 12. Following \cite{pivnets} and \cite{raftpiv}, we use endpoint error (EPE) as the primary evaluation metric for comparing different algorithms. The EPE is the spatially mean value of Euclidean distance between the predicted flow $\mathbf{u}_\text{pred}$ and its ground truth $\mathbf{u}_\text{gt}$.

\begin{table*}[ht]
\centering
\small
\begin{tabular}{l|cccccccc|cc}
\hline
\multirow{2}{*}{Methods} & \multicolumn{2}{c}{Channel} & \multicolumn{2}{c}{Isotropic} & \multicolumn{2}{c}{Mhd} & \multicolumn{2}{c|}{Mixing} & \multicolumn{2}{c}{Average} \\
\cline{2-3} \cline{4-5} \cline{6-7} \cline{8-9} \cline{10-11} 
& $\Delta t$=21 & $\Delta t$=11 & $\Delta t$=21 & $\Delta t$=11 & $\Delta t$=21 & $\Delta t$=11 & $\Delta t$=21 & $\Delta t$=11 & $\Delta t$=21 & $\Delta t$=11 \\
\hline
(A) coarse-to-fine HS & 6.790 & 0.509 & 7.415 & 1.998 & 7.683 & 2.558 & 9.388 & 1.954 & 7.819 & 1.755 \\
(B) RAFT-PIV-Image           & 0.377          & 0.215          & 2.181          & 1.252          & 1.336          & 0.842          & 1.875          & 1.064          & 1.442          & 0.843 \\
  (C) RAFT-PIV-Spike & 0.347 & 0.208 & 1.223 & 0.755 & 0.857 & 0.557 & 0.958 & 0.581 & 0.846 & 0.525 \\
 (D) PIV-NetS-Image & 0.545 & 0.301 & 2.560 & 1.497 & 1.574 & 0.985 & 2.247 & 1.320 & 1.731 & 1.026 \\
 (E) PIV-NetS-Spike & 0.553 & 0.289 & 1.507 & 0.874 & 1.252 & 0.734 & 1.382 & 0.782 & 1.173 & 0.670 \\
  (F) SCFlow & 0.354 & 0.222 & 2.232 & 1.305 & 1.374 & 0.901 & 2.132 & 1.169 & 1.522 & 0.899 \\
  (G) HiST-SFlow & 0.369 & 0.224 & 1.271 & 0.796 & 0.964 & 0.611 & 1.029 & 0.637 & 0.908 & 0.567 \\
  (H) Flowformer-Spike & 0.286 & \textbf{0.147} & 1.069 & 0.642 & 0.727 & 0.437 & 0.806 & 0.472 & 0.722 & 0.425 \\
  (I) SIV (ours)       & \textbf{0.239} & 0.159 & \textbf{0.896} & \textbf{0.587} & \textbf{0.613} & \textbf{0.422} & \textbf{0.680} & \textbf{0.441} & \textbf{0.607} & \textbf{0.402} \\
\hline
\end{tabular}
\caption{EPE ($\downarrow$) results on PSSD dataset - Problem 1. The best results are
marked in bold.}
\label{tab:problem1}
\end{table*}

\begin{table*}[ht]
\centering
\small
\begin{tabular}{l|cccccccc|cc}
\hline
\multirow{2}{*}{Methods} & \multicolumn{2}{c}{Channel} & \multicolumn{2}{c}{Isotropic} & \multicolumn{2}{c}{Mhd} & \multicolumn{2}{c|}{Mixing} & \multicolumn{2}{c}{Average} \\
\cline{2-3} \cline{4-5} \cline{6-7} \cline{8-9} \cline{10-11} 
& $\Delta t$=21 & $\Delta t$=11 & $\Delta t$=21 & $\Delta t$=11 & $\Delta t$=21 & $\Delta t$=11 & $\Delta t$=21 & $\Delta t$=11 & $\Delta t$=21 & $\Delta t$=11 \\
\hline
  (A) coarse-to-fine HS & 20.691 & 8.244 & 17.164 & 8.051 & 17.469 & 8.170 & 17.131 & 7.936 & 18.114 & 8.100 \\

  (B) RAFT-PIV-Image & 0.665 & 0.368 & 4.730 & 2.635 & 2.320 & 1.366 & 4.305 & 2.368 & 3.005 & 1.684 \\
  
  (C) RAFT-PIV-Spike & 0.688 & 0.412 & 2.410 & 1.483 & 1.468 & 0.959 & 2.071 & 1.241 & 1.659 & 1.024 \\
  
  (D) PIV-NetS-Image & 1.163 & 0.683 & 4.866 & 2.830 & 2.673 & 1.643 & 4.347 & 2.500 & 3.262 & 1.914 \\

  (E) PIV-NetS-Spike & 0.887 & 0.540 & 3.172 & 1.818 & 2.392 & 1.410 & 3.067 & 1.744 & 2.379 & 1.378 \\

  (F) SCFlow & 0.694 & 0.403 & 4.689 & 2.613 & 2.387 & 1.391 & 4.326 & 2.380 & 3.024 & 1.696 \\

(G) HiST-SFlow & 0.556 & 0.364 & 2.212 & 1.365 & 1.404 & 0.998 & 1.911 & 1.179 & 1.521 & 0.976 \\
  (H) Flowformer-Spike & 0.533 & 0.284 & 1.975 & 1.179 & 1.214 & 0.748 & 1.668 & 0.972 & 1.348 & 0.796 \\

  (I) SIV (ours) & \textbf{0.370} & \textbf{0.254}& \textbf{1.687} & \textbf{1.080} & \textbf{1.071} & \textbf{0.722}& \textbf{1.449}& \textbf{0.912}& \textbf{1.144}& \textbf{0.742}\\
\hline
\end{tabular}
\caption{EPE ($\downarrow$) results on PSSD dataset - Problem 2. The best results are
marked in bold.}
\label{tab:problem2}
\end{table*}

\begin{table*}[ht]
\centering
\small
\begin{tabular}{l|cccccccc|cc}
\hline
\multirow{2}{*}{Methods} & \multicolumn{2}{c}{Channel} & \multicolumn{2}{c}{Isotropic} & \multicolumn{2}{c}{Mhd} & \multicolumn{2}{c|}{Mixing} & \multicolumn{2}{c}{Average} \\
\cline{2-3} \cline{4-5} \cline{6-7} \cline{8-9} \cline{10-11} 
& $\Delta t$=21 & $\Delta t$=11 & $\Delta t$=21 & $\Delta t$=11 & $\Delta t$=21 & $\Delta t$=11 & $\Delta t$=21 & $\Delta t$=11 & $\Delta t$=21 & $\Delta t$=11 \\
\hline

(A) coarse-to-fine HS & 7.908 & 1.044 & 7.924 & 2.706 & 8.345 & 3.130 & 7.854 & 2.716 & 8.008 & 2.399 \\
   (B) RAFT-PIV-Image & 0.447& 0.244 & 2.694 & 1.495 & 1.447 & 0.888 & 2.311 & 1.271 & 1.725  & 0.974 \\

  (C) RAFT-PIV-Spike & 0.320 & 0.200 & 1.203 & 0.740 & 0.798 & 0.530 & 0.946 & 0.576 & 0.817 & 0.512 \\
  
  (D) PIV-NetS-Image & 0.667 & 0.343 & 2.786 & 1.602 & 1.641 & 0.956 & 2.393 & 1.378 & 1.872 & 1.070 \\

  (E) PIV-NetS-Spike & 0.476 & 0.247 & 1.420 & 0.856 & 1.157 & 0.716 & 1.342 & 0.774 & 1.099 & 0.648 \\

  (F) SCFlow & 0.462 & 0.269 & 1.850  & 0.837  & 1.300 & 0.774 & 1.972 & 0.769 & 1.396 & 0.662 \\

  (G) HiST-SFlow & 0.329 & 0.180 & 1.179 & 0.640 & 0.832 & 0.452 & 0.960 & 0.476 & 0.825 & 0.437 \\

  (H) Flowformer-Spike & 0.269 & \textbf{0.145} & 1.052 & 0.620 & 0.716 & 0.429 & 0.813 & 0.481 & 0.713 & 0.419 \\

(I) SIV (ours) & \textbf{0.234}& 0.155 & \textbf{0.880}& \textbf{0.573} & \textbf{0.598} & \textbf{0.404} & \textbf{0.677} & \textbf{0.438} & \textbf{0.598} & \textbf{0.392}\\
\hline
\end{tabular}
\caption{EPE ($\downarrow$) results on PSSD dataset - Problem 3. The best results are
marked in bold.}
\label{tab:problem3}
\end{table*}

\subsection{ Comparative Experiments}
We compare our method with several open-source baselines, including classical PIV algorithms \cite{cai2018motion}, image-based PIV networks (PIV-NetS \cite{pivnets}, RAFT-PIV \cite{raftpiv}), spike-adapted optical flow networks (e.g., Flowformer with spike input \cite{flowformer}), and spike-based optical flow networks (SCFlow \cite{hu2022optical}, HiST-SFlow \cite{hist}). All networks trained on spikes (incl. (C)(E)(F)(G)(H)(I) in Table \ref{tab:problem1}, \ref{tab:problem2}, and \ref{tab:problem3}) use \textbf{voxel-grid inputs} to ensure a fair evaluation. All image baselines (incl. (A)(B)(D) in Table \ref{tab:problem1}, \ref{tab:problem2}, and \ref{tab:problem3}) use sharp frames containing only artificially added noise (as shown in Figure \ref{fig:piv_scene}, \textbf{not spike reconstructions}) simulating high-speed cameras to highlight the differences between image and spike inputs. The image is temporally aligned with the intermediate moments of the spike stream.

As shown in Tables \ref{tab:problem1}, \ref{tab:problem2}, and \ref{tab:problem3}, $\Delta t = 21$ and $\Delta t = 11$ represent the time intervals between two spike streams used for velocity field prediction. Our SIV method achieves the lowest average EPE across all sub-datasets. Compared to baseline approaches, SIV demonstrates superior prediction accuracy in both low-velocity (Problem 1) and high-velocity (Problem 2) scenarios. Additionally, in high dynamic range conditions (Problem 3), SIV effectively leverages the advantages of spike-based input modalities, addressing the fundamental limitations of conventional frame-based systems.

Figure \ref{visualization} visualizes the predictions on three problems. The top row visualizes the velocity fields using the standard optical flow color coding \cite{colorcoding}, and the bottom row shows the EPE. Since shear layers are crucial for turbulence generation and development, it is noteworthy that SIV attains the lowest prediction error in both shear layers and small-scale vortices, while avoiding noticeable blurring, distortion, or discontinuities.

\subsection{Ablation Study}
\noindent \textbf{Ablation of modules.} All values in Table \ref{tab:ablation} and \ref{tab:ablation2} are the arithmetic mean across scenes ($\Delta t=21$), with the best results highlighted in bold. The results emphasize the effectiveness of DPHT, GE, MSVR, and $L_\text{grad}$. In Table \ref{tab:ablation}, comparisons between experiments (A) and (B) show the effectiveness of DPHT, while comparisons between (B) and (C), and (D) and (E), validate the contribution of the Graph Encoder (GE). Additionally, comparisons between (B) and (D), and (C) and (E), highlight the impact of MSVR.\\

\noindent \textbf{Ablation of $\beta$.} Table \ref{tab:ablation2} summarizes the ablation study on the weighting coefficient $\beta$ in the proposed loss function. The results show that $\beta = 0.3$ yields the best performance for the current model and dataset.

\begin{table}[ht]  
\centering
\setlength{\tabcolsep}{3pt}  
\renewcommand{\arraystretch}{1.2}  
\small  
\begin{tabular}{ccccccc}
  \hline
  \multirow{2}{*}{\#} & \multicolumn{3}{c}{Settings} &{Problem 1} &{Problem 2} & {Problem 3} \\
  \cline{2-4} 
  & DPHT & GE & MSVR & $\Delta t$=21 & $\Delta t$=21 &  $\Delta t$=21 \\
  \hline
  (A) & $\times$ & $\times$ & $\times$ & 0.805 & 1.450 & 0.776\\
  (B) & $\checkmark$ & $\times$ & $\times$  & 0.701 & 1.299 & 0.659 \\
  (C) & $\checkmark$ & $\checkmark$ & $\times$ & 0.625 & 1.167 & 0.616 \\
  (D) & $\checkmark$ & $\times$ & $\checkmark$ & 0.644 & 1.268 & 0.635 \\
  (E) & $\checkmark$ & $\checkmark$ & $\checkmark$ & \textbf{0.607} & \textbf{1.144} & \textbf{0.598} \\
  \hline
\end{tabular}
\caption{Ablation study of proposed modules, showing results for EPE ($\downarrow$). $\checkmark$ indicates module presence, $\times$ indicates absence. Setting (E) is our final model.}
\label{tab:ablation}
\end{table}

\begin{table}[ht]
\centering
\small  
\begin{tabular}{c|c|ccc} 
\hline
\# & $\beta$ & Problem 1 & Problem 2 & Problem 3 \\ 
\hline
(A) & 0.0 & 0.620 & 1.169 & 0.615 \\
(B) & 0.1 & 0.617 & 1.163 & 0.605 \\
(C) & 0.2 & 0.615 & 1.157 & 0.603 \\
(D) & 0.3 & \textbf{0.607} & 1.144 & \textbf{0.598} \\
(E) & 0.4 & 0.613 & 1.146 & 0.610 \\
(F) & 0.5 & 0.619 & \textbf{1.141} & 0.617 \\
\hline
\end{tabular}
\caption{Ablation study of $\beta$, showing results for EPE (↓).}
\label{tab:ablation2}
\end{table}
\section{Conclusion}
We propose a novel approach for fluid motion estimation using spike cameras and introduce a specialized network architecture, Spike Imaging Velocimetry (SIV), designed to address PIV tasks while adapting to the specific characteristics of fluid scenes. We develop the Detail-Preserving Hierarchical Transform (DPHT) module to preserve particle signal information during spike stream preprocessing, a Graph Encoder (GE) to extract contextual features from highly unstructured and turbulent flows, and a Multi-scale Velocity Refinement (MSVR) module to refine the velocity field. We also introduce the challenging and realistic Particle Scenes with Spike and Displacement (PSSD) dataset. Experimental results demonstrate that SIV achieves state-of-the-art performance on the PSSD dataset, accurately estimating velocities in small-scale yet critical flow structures such as shear layers and vortices. In summary, this study highlights the potential of spike cameras for fluid velocity measurement and offers new perspectives and directions for PIV network design.

This paper aims to demonstrate that spike cameras are a promising tool for PIV estimation. However, given the complexity of fluid flows, there remains substantial room for improvement: (i) although we have generated a large dataset, its scale and diversity should be further increased; (ii) noise should be modeled more comprehensively to achieve stronger performance on experimental data; and (iii) improving inference efficiency to enable real-time prediction would benefit both laboratory and industrial applications.

\section{Acknowledgements}
This work is supported by the Postgraduate Research \& Practice Innovation Program of Jiangsu Province (No. KYCX25\_0173), the National Natural Science Foundation of China under Grants 62371006, 62302016,  U24B20140, the Jiangsu Association for Science and Technology (JAST) Young Elite Scientists Sponsorship Program (No. JSTJ-2025-730), and the Beijing Natural Science Foundation (No. 3242008).

\bibliography{aaai2026}

\end{document}